\documentclass{article} 
\usepackage{iclr2021_conference,times}

\usepackage{amsmath,amsfonts,bm}

\def\eqref#1{equation~\ref{#1}}

\def\1{\bm{1}}

\DeclareMathAlphabet{\mathsfit}{\encodingdefault}{\sfdefault}{m}{sl}
\SetMathAlphabet{\mathsfit}{bold}{\encodingdefault}{\sfdefault}{bx}{n}

\DeclareMathOperator*{\argmax}{arg\,max}

\usepackage[utf8]{inputenc} 
\usepackage[T1]{fontenc}    
\usepackage{hyperref}       
\usepackage{url}            
\usepackage{booktabs}       
\usepackage{amsfonts}       
\usepackage{nicefrac}       
\usepackage{microtype}      
\usepackage{amsthm}
\usepackage{slashbox}
\usepackage{wrapfig}
\theoremstyle{definition}
\newtheorem{definition}{Definition}[section]

\usepackage{graphicx}
\usepackage{subcaption}
\usepackage{eqnarray}
\usepackage{amsmath,amsthm}
\usepackage{amssymb}
\usepackage{mathtools}
\usepackage{multirow}
\usepackage[ruled,vlined]{algorithm2e}
\DeclareMathOperator\erfinv{erfinv}

\usepackage{hyperref}
\usepackage{xcolor,colortbl}
\usepackage{color}

\title{Fine-Grained $\epsilon$-Margin Closed-Form \\Stabilization of Parametric Hawkes Processes}

\author{%
  Rafael Lima\\
  Samsung R\&D Institute Brazil\\
  Av Cambacicas, 1200, Campinas-SP, Brazil \\
  \texttt{rafael.goncalves.lima@gmail.com} \\
}
\iclrfinalcopy 
\begin{document}

\maketitle

\begin{abstract}
Hawkes Processes have undergone increasing popularity as default tools for modeling self- and mutually exciting interactions of discrete events in continuous-time event streams. A Maximum Likelihood Estimation (MLE) unconstrained optimization procedure over parametrically assumed forms of the triggering kernels of the corresponding intensity function are a widespread cost-effective modeling strategy, particularly suitable for data with few and/or short sequences. However, the MLE optimization lacks guarantees, except for strong assumptions on the parameters of the triggering kernels, and may lead to instability of the resulting parameters . In the present work, we show how a simple stabilization procedure improves the performance of the MLE optimization without these overly restrictive assumptions. This stabilized version of the MLE is shown to outperform traditional methods over sequences of several different lengths.
\end{abstract}

\section{Introduction}

Temporal Point Processes are mathematical objects for modeling occurrences of discrete events in continuous-time event streams. Hawkes Processes (HP) (\cite{AH71}), a particular form of point process, are particularly suited for capturing self- and mutual-excitation among events, i.e., when the arrival of one event makes future events more likely. HPs have recently undergone increasing popularization in both natural and social domains, such as Finance (\cite{EB15,JE16,BM12}), Social Networks (\cite{MR18,QZ15}), Earthquake Occurrences (\cite{YO81,AH02}), Criminal Activities (\cite{GM12,SL14}), Epidemiology (\cite{MR182}) and Paper Citation counts (\cite{HZ16}).

In HPs, the self- and mutual-excitation is represented by the addition of a term in the expected rate of event arrivals: the \textit{triggering kernel}. For datasets with few and/or short sequences, which are the case for many real-world applications, a parametric choice of the triggering function is the default modeling choice (\cite{JC20}), as a way of making use of domain knowledge to meaningfully capture the main aspects of the triggering effect. Further related HP works are discussed in the Appendix.

After the choice of the function, an unconstrained optimization is carried out for obtaining the MLE parameters. Unfortunately, both the non-convexity of the implicitly defined likelihood function, as well as eventual ill-conditioning of the initialization of the to-be-optimized self-triggering function parameters,may lead to a consequent instability of the resulting model, which is to say that the resulting HP may asymptotically result in an infinite number of events arriving in a finite time interval.

In the following sections, we:
\begin{enumerate}
\item Describe the MLE for Parametric HPs, along with the proposed choices for the triggering function;
\item Introduce the Fine-Grained Stabilization MLE procedure (FGS-MLE) as a way of improving the stability and reliability of the parametric HPs obtained through the MLE computation;
\item Validate the stabilization procedure on sets of event sequences of several different lengths, also accounting for misspecification of each parametric form, with respect to one another.
\end{enumerate}

\section{Parametric Maximum Likelihood Estimation of Hawkes Processes}
\label{sec: param_hp}

In the following, we restrict ourselves to the Univariate HP case. A realization of a Temporal Point Process over $[0,T]$ may be represented by a Counting Process $N_t$ such that $dN_t = 1$, if there is an event at time $t$, or $dN_t = 0$, otherwise. Alternatively, it may be represented by a vector $\mathcal{S} = \{t_1, t_2, ..., t_N\}$, which contains all the time coordinates of the event arrivals considered in the realization of the process.

A Conditional Intensity Function (CIF) $\lambda (t)$, may be defined over $N_t$ as $\lambda (t) dt= \mathbb{E} \{ dN_t = 1 | \mathcal{H}_t\}$, where $\mathcal{H}_t = \{ \forall t_i \in \mathcal{S} | t_i < t \}$.

The CIF of a Hawkes Process is defined as: $\lambda_{HP} (t)= \mu + \sum_{t_i < t} \phi(t-t_i)$, in which $\mu \in \mathbb{R^{+}}$ is the \textit{background rate} or \textit{exogenous intensity}, here taken as a constant, while $\phi (t) \in \mathbb{R^{+}}$ is the \textit{excitation function} or \textit{triggering kernel} (TK), which captures the influence of past events in the current value of the CIF.

In the present work, we will deal with the following five types of parametric TKs:

\textbf{(Exponential)} $EXP(\alpha,\beta) = \alpha e^{-\beta t}$ \hspace{1cm} \textbf{(Power-Law)} $PWL(K,c,p) = K(t+c)^{-p}$
\textbf{(Rayleigh)} $RAY(\gamma,\eta) = \gamma t e^{-\eta t^2}$ \hspace{2cm} \textbf{(Gaussian)} $GSS(\kappa,\tau, \sigma) = \kappa \mathrm{e}^{-\dfrac{(t - \tau)^2}{\sigma}}$

\textbf{(Tsallis Q-Exponential)} \hspace{5cm} $\text{$\alpha$,$\beta$,$\gamma$,$\eta$,K,c,$\kappa$,$\tau$,$\sigma$,a,q} > 0$ \text{  } $p > 1$
    \begin{equation}
    QEXP(a,q) = 
    \left\{
    \begin{array}{ll}
    a e^{-t} & \quad q = 1 \\
    a \left[ 1+ (q-1)t\right]^\frac{1}{(1-q)} & \quad q \neq 0 \text{ and } 1+ (1-q)t > 0 \\
    0 & q \neq 0 \text{ and } 1+ (1-q)t\leq 0
    \end{array}
    \right.
    \end{equation}

The Maximum Likelihood Estimation procedure for parametric HPs consists in, given a choice of triggering kernels, along with its implicitly defined form of CIF, finding a tuple of optimal parameters $(\boldsymbol{\mu}^{*},\theta^{*})$ such that:$(\boldsymbol{\mu}^{*},\theta^{*}) = \argmax_{\mu \in \mathbb{R}^{+}, \theta \in \Omega} \mathcal{L}_{\mathcal{S}}(\boldsymbol{\mu}, \theta)$,
where the loglikelihood function $\mathcal{L}_{\mathcal{S}}(\boldsymbol{\mu}, \boldsymbol{\theta})$ over a sequence $\mathcal{S}$ is defined by: $\mathcal{L}_{\mathcal{S}}(\boldsymbol{\mu}, \boldsymbol{\theta}) = \sum_{i \in N }\log \lambda_{HP} (t_i) - \int_{0}^{T} \lambda_{HP} (t) dt$.
In practice, this optimization is done through gradient-based updates, with a learning rate $\delta$, over $\mu$ and each of the parameters:
\begin{equation}
\mu^{i+1} = \mu^i + \delta \nabla_{\mu^i} \mathcal{L}_{\mathcal{S}}(\mu^i, \boldsymbol{\theta^i}) \;\;\;\;\;\;\;  \theta^{i+1} = \theta^i + \delta \nabla_{\theta^i} \mathcal{L}_{\mathcal{S}}(\mu^i, \boldsymbol{\theta^i})
\end{equation}
 up to a last iterate I, which is implicitly defined by a stopping criteria (e.g., Optimality Tolerance or Step Tolerance). Gradient-free methods' updates (e.g., Nelder-Mead) may be used analogously.

This procedure, first introduced in (\cite{TO79}), lacks theoretical guarantees with respect to the convexity of  $\mathcal{L}_{\mathcal{S}}(\boldsymbol{\mu}, \boldsymbol{\theta})$, over finite sequences, for a more general choice of $\phi(t)$. In practice, very restrictive assumptions are made, such as fixing $\alpha = \beta$ in the EXP($\alpha$,$\beta$) kernel, and constraining the initial values of the parameters, in advance, to be very close to the optimal values. Both of these are unfeasible for practical applications, when the true values of the parameters are hardly known. Prescinding from these often lead to the resulting final values of parameters to be very far from the optimal, and frequently corresponding to unstable configurations, i.e., when $|\phi| = \int_{0^-}^{\infty} \phi (t) dt > 1$.
This condition leads the underlying HP to be asymptotically unstable, generating an infinite number of events over a finite time interval as the length of the sequence $\mathcal{S}$ extends up to infinity.

In the  next section, we will present a stabilization method, which preserves the stability of the corresponding $\phi (t)$ for each of the five introduced parametric forms of triggering kernels, without constraining their originally considered parameter space $\Omega$. This stabilization is performed after the last iterate value of the parameters, through closed-form computations.

\section{Fine-Grained $\epsilon$-Margin Closed-Form Stabilization of Parametric Hawkes Processes}

\begin{figure}
\centering
\includegraphics[width=0.8\linewidth]{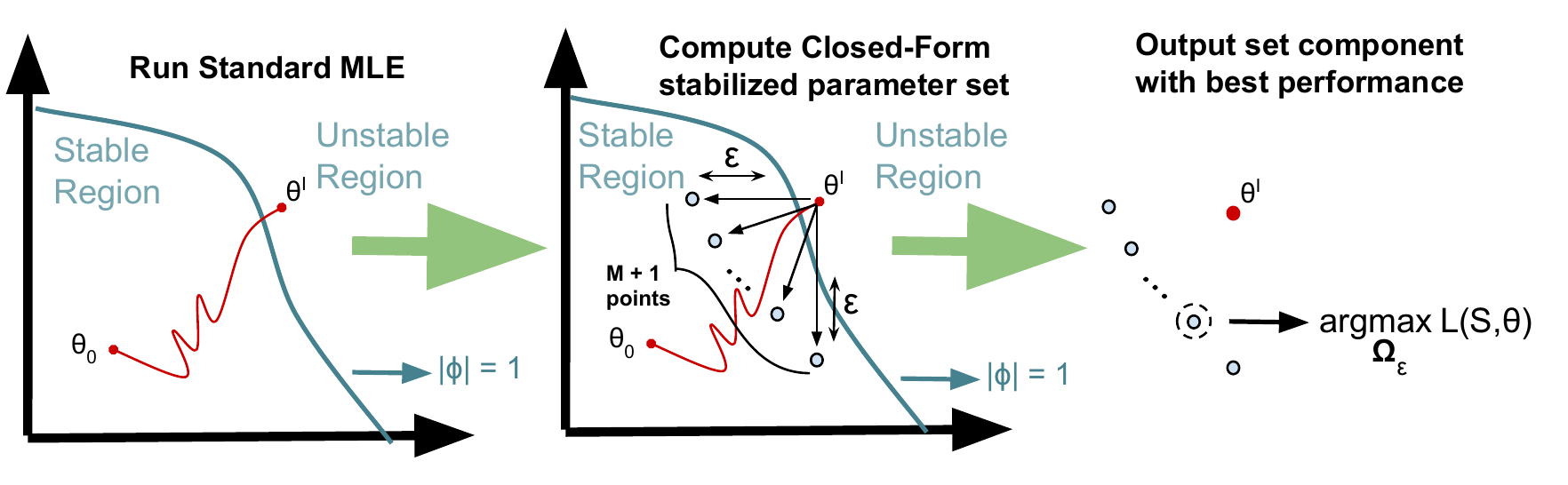}
\caption{{\footnotesize Diagram of the two-dimensional stabilization procedure. Given values for M and $\epsilon$, the algorithm consists in finding (M+1) points inside the stable region for the two-dimensional parameter vector $\theta$, each at a distance $\epsilon$ from the boundary. The (M+1) points represent a gradual change of the relevance of each component of $\theta$ to the its final stabilized value. The procedure for higher-dimensional $\theta$ follows analogously.}}
\label{fig:stab}
\end{figure}

In this section, we describe the stabilization procedure for the MLE of parametric HPs. It essentially consists of finding stable configurations for both $\mu$ and $\phi(t)$ in a closed-form way, with a pre-defined values for stability margin and resolution. An intuitive graphical description of the stabilization method is shown in Figure \ref{fig:stab}

\theoremstyle{definition}
\begin{definition}{\textbf{Stabilized Hawkes Process}}\label{def: stabhp}

Given a Hawkes Process implicitly defined by its background rate $\mu \in \mathbb{R}^{+}$ and its triggering kernel $\phi (t): \Omega \times \mathbb{R}^{+} \rightarrow \mathbb{R}^{+}$, its corresponding Stabilized Hawkes Process is found by obtaining new parameter values $\mu_{\epsilon} \in \mathbb{R}_{\epsilon}^{+}$ and $\phi_{\epsilon} (t): \Omega_\epsilon \times \mathbb{R}^{+} \rightarrow \mathbb{R}^{+}$ such that:
\begin{equation}
|\phi_\epsilon| = \int_{0^-}^{\infty} \phi_{\epsilon} (t) = \dfrac{1}{1+\epsilon}, \;\;\;\; \mu_\epsilon = \dfrac{\Lambda}{1 + |\phi_\epsilon|} \;\;\;\; \rightarrow \;\;\;\; \boldsymbol{\theta_\epsilon} = \argmax_{\theta \in \Omega_\epsilon} \mathcal{L}_{\mathcal{S}}(\mu_\epsilon, \boldsymbol{\theta})
\end{equation}
where $\Omega_\epsilon$ is a ($M \times \dim(\theta)$)-sized set defined in Definition \ref{def: omegaeps}
\end{definition}

\theoremstyle{definition}
\begin{definition}{\textbf{$\epsilon$-Margin Fine-Grained Parameter Set $\Omega_\epsilon$}}\label{def: omegaeps}

Given last iterate parameter vector $\theta^I \in \Omega$ (with $D = \dim(\Omega)$), a margin parameter $\epsilon \in (0,1)$, a Stabilization Resolution parameter $M \geq 3$, and two subindexes, j and k ($j \neq k$), from the $\theta^I$ parameter vector, the $\epsilon$-Margin Fine-Grained Parameter Set $\Omega_\epsilon$ is defined as a $(M+1) \times \dim(\Omega)$-sized tuple such that $\Omega_\epsilon = \{\theta_I, \theta_0, \theta_1, ..., \theta_M\}$, with each $\theta_i = \{\theta_i^d\}_{d=1}^{D}$ $(i \in \{0,1, ..., M\})$ computed such that, given $|\phi| =\int_{0^-}^{\infty} \phi (t) dt$ computed using the kernel parameter values from $\theta^I$:
\begin{equation}
\int_{0^-}^{\infty} \phi_{\theta_i^j} (t) dt = |\phi|^{\dfrac{M-i}{M}} \slash (1+\epsilon)^{\dfrac{i}{M}} \;\;\;\;\;\;\; \int_{0^-}^{\infty} \phi_{\theta_i^k} (t) dt = |\phi|^{\dfrac{i}{M}} \slash (1+\epsilon)^{\dfrac{M-i}{M}},
\label{eq: renorm}
\end{equation}
where $\phi_{\theta_i^j} (t)$ denotes the original $\phi (t)$ with only the j-th parameter replaced by a value which leads $\phi (t)$ to satisfy the given stabilization condition.  Details on the derivation of closed-form expressions for the set $\Omega_{\epsilon}$, computed for each of the five parametric kernels, are discussed in the Appendix.
\end{definition}

\vspace{-0.3cm}
\section{Experiments}


For an initial empirical validation of the proposed stabilization strategy, we performed experiments with synthetic data generated from each of the five parametric forms. Since last iterate $\theta^I \in \Omega_\epsilon$, then clearly $\max_{\theta \in \Omega_\epsilon} \mathcal{L} (\theta, \mathcal{S}) \geq \mathcal{L} (\theta^I,\mathcal{S})$. Our goal is, then, verifying the rate with which the stabilization algorithm output strictly outperforms $\theta^I$, including those cases of model misassignment (e.g., an exponential kernel is used to fit a sequence generated by a gaussian kernel, and so on). 

\begin{wrapfigure}{r}{0.6\textwidth}
\begin{minipage}[c][8.2cm][t]{.5\textwidth}
\vspace*{\fill}
\centering
\includegraphics[width=1.2\linewidth]{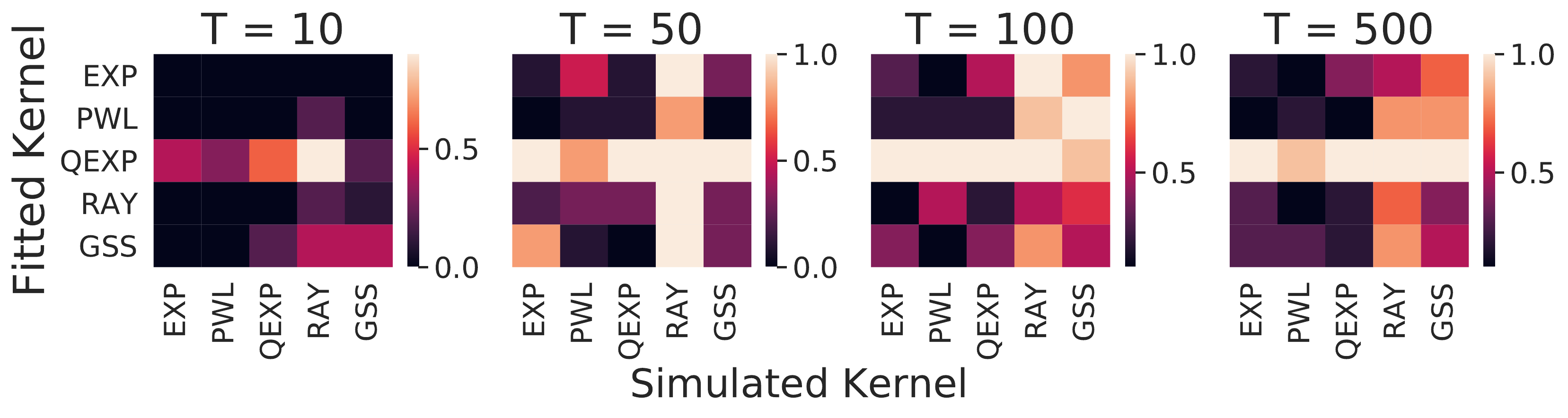}
\subcaption{Influence of T.}
\label{fig:T_influence}
\includegraphics[width=1.2\linewidth]{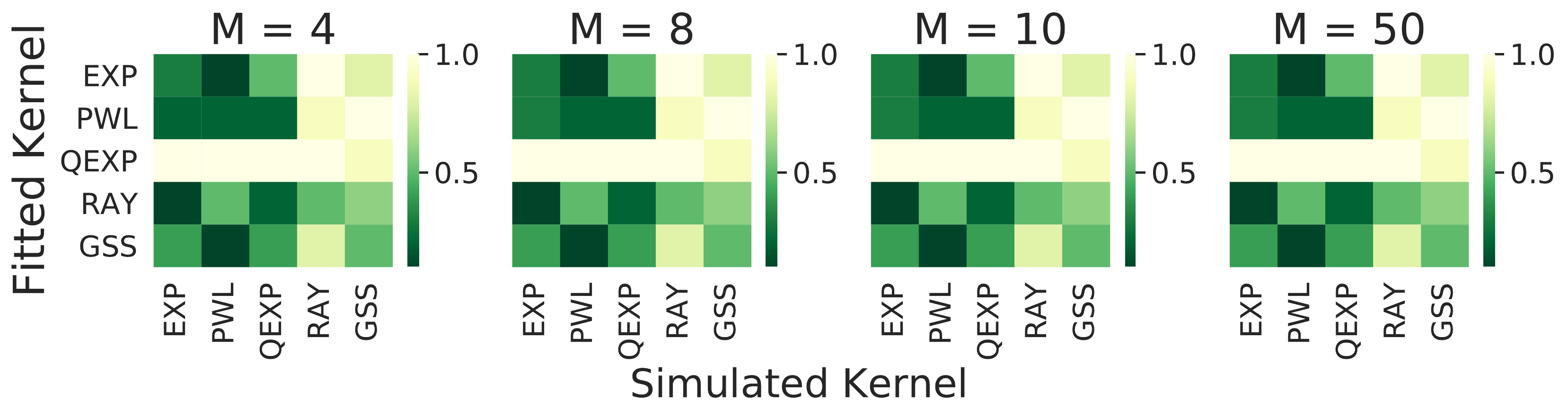}
\subcaption{Influence of M.}
\label{fig:M_influence}
\includegraphics[width=1.2\linewidth]{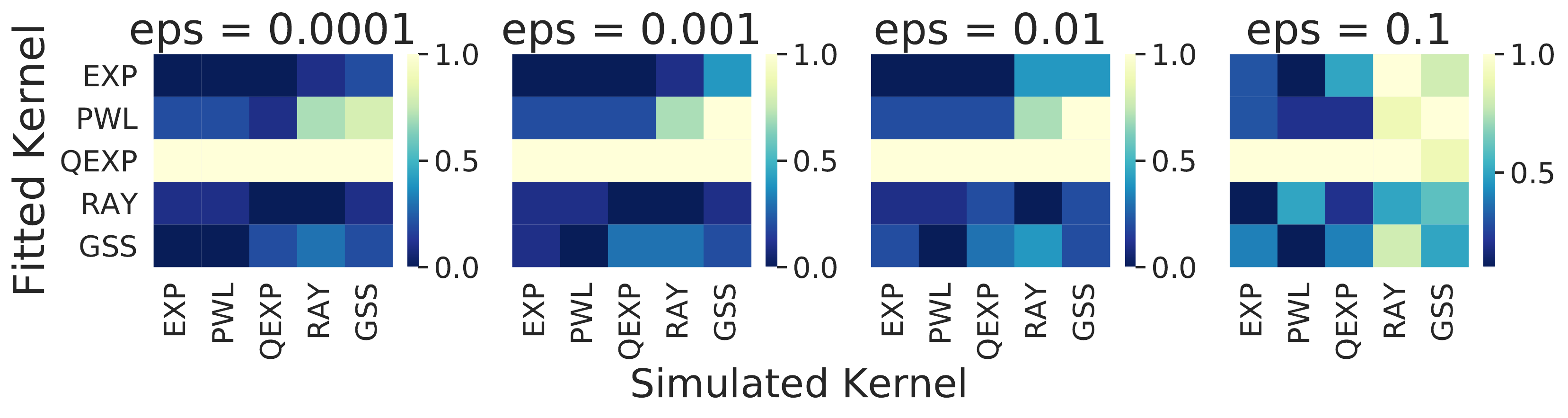}
\subcaption{Influence of $\epsilon$.}
\label{fig:eps_influence}
\end{minipage}
\caption{Influence of different parameters on the performance of the performance of the MLE-STAB algorithm.}
\label{fig:influence}
\end{wrapfigure}

In the following, we investigate the performance of the stabilization algorithm over sequences of several different horizons T, and also verify the influence of margin parameter $\epsilon$ and stabilization resolution M on the success rate of stabilized kernels. The results are shown in Figure \ref{fig:influence}. The MLE optimization procedure is carried out with the standard Nelder-Mead method implementation from the Python \textit{SciPy} library\footnote{Experiments with other optimization methods are discussed in the Appendix.}.  As an example, a success rate of 0.9 for EXP kernel fitted over sequences generated with GSS kernels means that the likelihood of the parameters returned by the stabilization method is strictly higher than the original MLE last iterate $\theta^I$ in 90 \% of the corresponding sequences. It can be observed that the MLE-STAB improves the performance of the model over a wide range of parameters, even in the case of kernel misassignment.

\subsection{Baseline Comparisons on Real and Synthetic Data}

For testing the quality of our stabilization algorithms, hereby denoted \textbf{MLE- STAB}, we performed experiments with real data. The chosen datasets were "Hawkes","Retweet" and "Missing". Their description and content can be found in (\cite{HM17}). The baselines chosen were those which allowed for a more flexible modeling of the triggering function. They are the implementations of \textbf{HawkesEM}, \textbf{HawkesBasisKernels} and \textbf{HawkesConditionalLaw}, publicly available in the \textit{tick} library (\cite{EB17}).

\begin{wraptable}{r}{8cm}
\centering
\begin{tabular}[t]{ccc} \toprule
    Dataset \slash Method & MLE-STAB & HawkesEM \\ \midrule
    Hawkes  & \textbf{5.30}  & 0.039 \\ 
    Missing  & \textbf{5.27}  & -0.06 \\ 
    Retweet  & \textbf{71.29}  & -20.68 \\ \bottomrule
\end{tabular}
\caption{{\footnotesize Comparison of the stabilization method with baseline over standardized datasets. The result shown corresponds to the average, across all sequences, of the loglikelihood of the method on the test portion of the given sequence (final $30\%$ of the event horizon $[0,T]$). The loglikelihood is normalized by the number of events contained in this given test portion.}}\label{tab:baseline_comp}
\end{wraptable} 

We intend to show here that, for most real-world applications, the number of events associated with a given self-triggering dynamics is usually small. This fact is advantageous to simple parametric functions, such as the ones we presented here. And the improvements proposed by the MLE-STAB algorithm works on improving the predictive power of these simpler parametric models. The models are fitted in the events up to $70\%$ of T, here taken as the time of the last event. And then have their loglikelihood evaluated in the final $30\%$. Unfortunately, the very small number of events was prohibitive for the \textbf{HawkesBasisKernels} and \textbf{HawkesConditionalLaw} models, thus only results for \textbf{HawkesEM} and \textbf{MLE-STAB} are displayed. The final average across all sequences of the dataset is shown in Table \ref{tab:baseline_comp}.


\section{Conclusions}

Hawkes Processes are particularly suited for modeling self- and mutually exciting interactions of discrete events in continuous-time event streams, what corroborates the increasing popularity they have enjoyed across many domains, such as Social Network analysis and High-Frequency Finance. A Maximum Likelihood Estimation (MLE) unconstrained optimization procedure over parametrically assumed forms of the triggering kernels of the corresponding intensity function are a widespread cost-effective modeling strategy, particularly suitable for data with few and/or short sequences. However, the MLE optimization lacks guarantees, except for strong assumptions on the parameters of the triggering kernels, and may lead to instability of the resulting parameters . In the present work, we show how a simple stabilization procedure improves the performance of the MLE optimization without these overly restrictive assumptions, while also accounting for model misassignment cases. This stabilized version of the MLE is shown to outperform the standard method over sequences of several different lengths. The effects of resolution and margin parameters are also evaluated, as well as baseline comparisons over standardized datasets.

\bibliography{references}
\bibliographystyle{iclr2021_conference}

\appendix
\section{Appendix}

\subsection{Closed-Form Computation of $\Omega_\epsilon$ for Parametric Hawkes Processes}

In this subsection, we detail the closed-form calculations of the stabilized parameter set $\Omega_\epsilon$ for each of the parametric forms for $\phi (t)$ detailed at Section \ref{sec: param_hp}.

\subsubsection{Example 1: Closed-Form Stabilization of EXP($\alpha$,$\beta$)}

For $\phi (t) = EXP (\alpha, \beta)$, we have $\theta^I = (\alpha^I, \beta^I) \in \mathbb{R}_{2}^{+}$. By defining M, $\epsilon$, j = 1 and k = 2, with $\alpha = \theta^1$, $\beta = \theta^2$ and $|\phi| = \int_{0^{-}}^{\infty} \phi (t)$, we can find tuples ($\alpha'$,$\beta'$) such that:
\begin{equation}
\dfrac{\alpha'}{\beta^I} = |\phi|^{\dfrac{M-i}{M}} \slash (1+\epsilon)^{\dfrac{i}{M}} \;\;\;\;\;\;\;\;\;\;\;\;\;\;\; \dfrac{\alpha^I}{\beta'} = |\phi|^{\dfrac{i}{M}} \slash (1+\epsilon)^{\dfrac{M-i}{M}}
\;\;\;\;\;\;\;\;\;\;\;\;\;\;\; \dfrac{\alpha^I}{\beta^I} = |\phi|,
\end{equation}
for each value of $i \in \{0, 1, ..., M\}$ as given in Equation \ref{eq: renorm}. From that, we can readily find:
\begin{equation}
\Omega_\epsilon = \{(\alpha^I,\beta^I)\} \cup \{(\alpha^I \slash (|\phi|(1+\epsilon))^{\dfrac{i}{M}}),\beta^I( |\phi|(1+\epsilon))^{\dfrac{M-i}{M}})\}_{i=0}^{M}
\end{equation}

\subsubsection{Example 2: Closed-Form Stabilization of PWL(K,c,p)}

For $\phi (t) = PWL (K, c, p)$, we have $\theta^I = (K^I, c^I, p^I) \in \mathbb{R}_{3}^{+}$. By defining M, $\epsilon$, j = 1 and k = 2, with $K = \theta^1$ and $c = \theta^2$, we can readily find:
\begin{equation}
\Omega_\epsilon = \{(K^I, c^I, p^I)\} \cup \{(K^I \slash (|\phi|(1+\epsilon))^{\dfrac{i}{M}},c^I((|\phi|(1+\epsilon))^{\dfrac{M-i}{M}})^{\frac{1}{p^I-1}},p^I)\}_{i=0}^{M} 
\end{equation}
 By defining j = 1 and k = 3, with $K = \theta^1$ and $p = \theta^3$, we can readily find:
\begin{equation}
\Omega_\epsilon = \{(K^I,c^I,p^I)\} \cup {(K^I \slash (|\phi|(1+\epsilon)), c^I, p^I)} \cup \{(K^I \slash (|\phi|(1+\epsilon))^{\dfrac{i}{M}},c^I,1+\frac{W(\Delta^i \log c^I)}{\log c^I})\}_{i=0}^{M-1} 
\end{equation}
where 
\begin{equation}
\Delta^i = (|\phi| (1+\epsilon))^{\dfrac{M-i}{M}} (p^I-1) {c^I}^{(p^I-1)}
\end{equation}
and W is a standard function, taken as the analytical continuation of branch 0 of the product log (Lambert-W) function.

\subsubsection{Example 3: Closed-Form Stabilization of QEXP(a,q)}

For $\phi (t) = QEXP(a, q)$, we have $\theta^I = (a^I, q^I) \in \mathbb{R}_{2}^{+}$. By defining M, $\epsilon$, j = 1 and k = 2, with $a = \theta^1$, $q = \theta^2$ and $|\phi| = \int_{0^{-}}^{\infty} \phi (t)$, we can readily find:
\begin{equation}
\Omega_\epsilon = \{(a^I,q^I)\} \cup \{(\frac{a}{(|\phi|(1+\epsilon)})^{\dfrac{i}{M}},2-(2-q)(|\phi|(1+\epsilon))^{\dfrac{M-i}{M}})\}_{i=0}^{M}
\end{equation}
\subsubsection{Example 4: Closed-Form Stabilization of RAY($\gamma$,$\eta$)}

For $\phi (t) = RAY(\gamma, \eta)$, we have $\theta^I = (\gamma^I, \eta^I) \in \mathbb{R}_{2}^{+}$. By defining M, $\epsilon$, j = 1 and k = 2, with $\gamma = \theta^1$, $\eta = \theta^2$ and $|\phi| = \int_{0^{-}}^{\infty} \phi (t)$, we can readily find:
\begin{equation}
\Omega_\epsilon = \{(\gamma^I,\eta^I)\} \cup \{(\gamma^I \slash (|\phi|(1+\epsilon))^{\dfrac{i}{M}}),\eta^I( |\phi|(1+\epsilon))^{\dfrac{M-i}{M}})\}_{i=0}^{M}
\end{equation}

\subsubsection{Example 5: Closed-Form Stabilization of GSS($\kappa$,$\tau$,$\sigma$)}

For $\phi (t) = GSS(\kappa,\tau,\sigma)$, we have $\theta^I = (\kappa^I, \tau^I, \sigma^I) \in \mathbb{R}_{3}^{+}$. By defining M, $\epsilon$, j = 1 and k = 2, with $\kappa = \theta^1$, $\tau = \theta^2$ and $|\phi| = \int_{0^{-}}^{\infty} \phi (t)$, we can readily find:
\begin{equation}
\begin{split}
\Omega_\epsilon = &\{(\kappa^I,\tau^I, \sigma^I)\} \cup \{(\kappa^I \slash (|\phi|(1+\epsilon))^{\dfrac{i}{M}},\tau^I \dfrac{\erfinv{(2 \slash (\kappa \sqrt{\pi \sigma}(1+\epsilon)^{\dfrac{M-i}{M}}) - 1})}{\erfinv{(2 |\phi|^{\dfrac{M-i}{M}} \slash (\kappa \sqrt{\pi} \sqrt{\sigma})-1})}, \sigma^I)\}_{i=0}^{M-1} \\ 
& \cup \{(\kappa^I \slash (|\phi|(1+\epsilon)), \tau^I, \sigma^I)\},
\end{split}
\end{equation}
where \textit{erfinv} is the inverse of the error function ($ erf (z) = \dfrac{2}{\sqrt{\pi}} \int_{0}^{z} \mathrm{e}^{-x^2} dx$). In practice, the arguments of \textit{erfinv} functions are constrained to be less than 1.

\subsection{Related Works}

The ubiquity of applicability of HPs has given rise to a myriad of modeling approaches for the triggering effect, inspired by parametric functions (e.g., Exponential, Power-Law, Gaussian) (\cite{TO79,HX16,YL182}) and Gaussian Processes (\cite{RZ192}). For data hungry applications, choices such as non-parametric methods (\cite{YY17, EB16, EL11,RZ19}), Neural Networks (NN) (\cite{HM17,ND16,SX18,TO19,JS19}), Attention Models (\cite{QZ19,SZ20}), have been proposed. Furthermore, improvements regarding real-world data limitations and challenges are developed in (\cite{CS18,WT19,FS19}).

In the parametric approach, several forms have been proposed, such as Exponential (\cite{AH71}), Power-Law (\cite{QZ15,EB162}), Rayleigh (\cite{RL18}), Tsallis Q-Exponential (\cite{RL18}) and Gaussian (\cite{HW20}). An additional, recently introduced, parametric modeling choice, the Mittag-Leffler function (\cite{JC20}), which is not feasible for explicit gradient-based optimization, is left for future work.

A recent work (\cite{NH19}) proposes a Convex Conjugate approach to tackle the non-convexity of the likelihood function. Regarding the stability of parameters, (\cite{DK13}) performs a theoretical analysis of unstable HPs, while (\cite{TJ15,EB12}) point to HPs with nearly unstable parameter as suitable for some application domains. Nevertheless, (\cite{TJ15}) restricts itself to the normalization of exponential kernel by its amplitude parameter, while (\cite{EB12}) concerns itself uniquely with asymptotic results, i.e., when the sequence of events is observed for a time window $[0,T]$, with $(T \rightarrow \infty)$. The work closest to ours is the one in (\cite{RL18}), which can be shown to be a particular case of our algorithm, for a specific value of the Stabilization Resolution (SR) parameter (M=2).

\subsection{Performance across different optimization methods}

For verifying the performance of the stabilization algorithms beyond the default choice of Nelder-Mead method, we computed the average loglikelihood improvement for each kernel type across the following optimization methods: \textbf{Nelder-Mead}, \textbf{Conjugate Gradient (CG)}, \textbf{BFGS}, \textbf{L-BFGS-B}, \textbf{TNC}, \textbf{COBYLA} and \textbf{SLSQP}. 

All the methods were run with the default \textit{SciPy} implementations, and the routines resulting in overflow were ignored. The parameters used were: T = 100, $\epsilon$ = 0.1 and M = 6. The results are shown in Table \ref{tab:opt_comp}. The stabilization method shows increase in performance across all methods but CG.

\begin{table}[ht]
\centering
\begin{tabular}[t]{cccccccc} \toprule
     Method & Nelder-Mead & CG & BFGS & L-BFGS-B & TNC & COBYLA & SLSQP \\ \midrule
    $\Delta \mathcal{L}$  & +520.86  & +0.0 & +131.83 & +124.57 & +202.65 & +682.47 & +109.45 \\ \bottomrule 
\end{tabular}
\caption{Comparison of the stabilization method among different optimization methods. $\Delta \mathcal{L}$ is the average difference in loglikelihood among the stabilized kernel and the original last iterate kernel across all sequences.}
\label{tab:opt_comp}
\end{table}%
The kernel types with largest increase in performance were EXP and QEXP.

\subsection{Experimental Reproducibility Guidelines}

For experiments with synthetic data, we performed experiments with sequences generated from each of the five parametric forms, with $\mu$ set as 0.5, and parameter values:
\begin{equation}
    EXP(1.0,1.1) \;\;\;\; PWL(0.9,1.0,2.0) \;\;\;\; QEXP(0.8,1.1) \;\;\;\; RAY(1.2,1.0) \;\;\;\; GSS(0.5,0.5,1.0),
\end{equation}

For the baseline comparison, initially, the chosen datasets were "Hawkes","Retweet" and "Missing", "Mimic" and "Meme". Their description and content can be found in (\cite{HM17}). For those datasets with more than one fold, only the first fold was used. Besides, only sequences with more than 20 events were considered. The "Meme" dataset had all sequences with a number of events which disallowed the fitting of all baselines, while "Mimic" dataset resulted in only one sequence being properly fitted for comparison. Thus, both these datasets were excluded from our comparison. After the model fitting, The loglikelihood of each sequence was then normalized by the number of events in the test portion. The performance of MLE-Stab is computed by choosing the parametric choice, out of the five possible ones, which performs best for each sequence.

\end{document}